\documentclass{article}
\usepackage{graphicx}
\usepackage{amssymb,amsmath}
\usepackage{multicol}
\usepackage{eqnarray}
\usepackage{caption}
\usepackage{subcaption}
\usepackage{textcomp}
\usepackage{multirow}
\usepackage{float}
\usepackage{bm}
\usepackage{color}
\usepackage{epstopdf}
\usepackage{hyperref}
\usepackage{geometry}
\usepackage{cite}
\begin{document}
\title{Neuromorphic Processing: A Unifying Tutorial} 
\date{} 
\author{Hamid Soleimani and Emmanuel.~M.~Drakakis}
\maketitle{} 
\begin{abstract}
All systolic or distributed neuromorphic architectures require power efficient processing nodes. In this paper, a unifying tutorial is presented which implements multiple neuromorphic processing elements using a systematic analog approach including synapse, neuron and astrocyte models. It is shown that the proposed approach can successfully synthesize multidimensional dynamical systems into analog circuitry with minimum effort.
\end{abstract}

\section{Introduction}
Given the application of the neuromorphic circuits, different approaches have been devised so far to mimic the biological behavior of the nervous system including special purpose computing architectures \cite{1,2,3,4,5}, digital \cite{6,7,8,9,10,11,12,13,14,15}, and analog platforms \cite{16,17,18,19,20,21,22,23,24}. Although all these approaches have their own merits and purposes, the interest upon mimicking biological behavior in their natural form (analog) has always been pursued rigorously by researchers. 

\par To this end, in this paper we apply a novel approach introduced in \cite{25} to systematically synthesize different neuromorphic processing elements into CMOS circuitry operating in strong--inversion. Same approach has been applied to log--domain circuits presented in \cite{16} with the purpose of achieving ultra--low power circuitry at the expense of slower time scales. In this tutorial, the application of the method is verified by synthesizing three different neuromorphic processing elements including synapse, neuron, and astrocyte models.  

\section{Method}
In this section, the novel approach proposed in \cite{25} is described which supports a systematic realization procedure of strong--inversion circuits capable of computing bilateral dynamical systems at higher speed compared to the previously proposed log--domain circuit \cite{16}. The current relationship of an NMOS and PMOS transistor operating in strong--inversion saturation when $\lvert V_{DS}\rvert>\lvert V_{GS}\rvert-\lvert V_{th}\rvert$ can be expressed as follows:
\begin{equation}\label{eq:51_1}
I_{D_n}=\frac{1}{2}\mu_n C_{ox}(\frac{W}{L})_n(V_{GS}-V_{th})^2
\end{equation}

\begin{equation}\label{eq:51_2}
I_{D_p}=\frac{1}{2}\mu_p C_{ox}(\frac{W}{L})_p(V_{SG}-V_{th})^2
\end{equation}

where $\mu_n$ and $\mu_p$ are the charge--carrier effective mobility for NMOS and PMOS transistors, respectively; $W$ is the gate width, $L$ is the gate length, $C_{ox}$ is the gate oxide capacitance per unit area and $V_{th}$ is the threshold voltage of the device.

\par Setting $k_{n}=\frac{1}{2}\mu_n C_{ox}(\frac{W}{L})_n$ and $k_{p}=\frac{1}{2}\mu_p C_{ox}(\frac{W}{L})_p$ in (\ref{eq:51_1}) and (\ref{eq:51_2}) and differentiating with respect to time, the current expression for $I_A$ (see Figure \ref{fig:51_1}) yields:

\begin{figure}[t]
\vspace{-20pt}
\normalsize
\centering
\includegraphics[trim = 0in 0in 0in 0in, clip, height=2.8in]{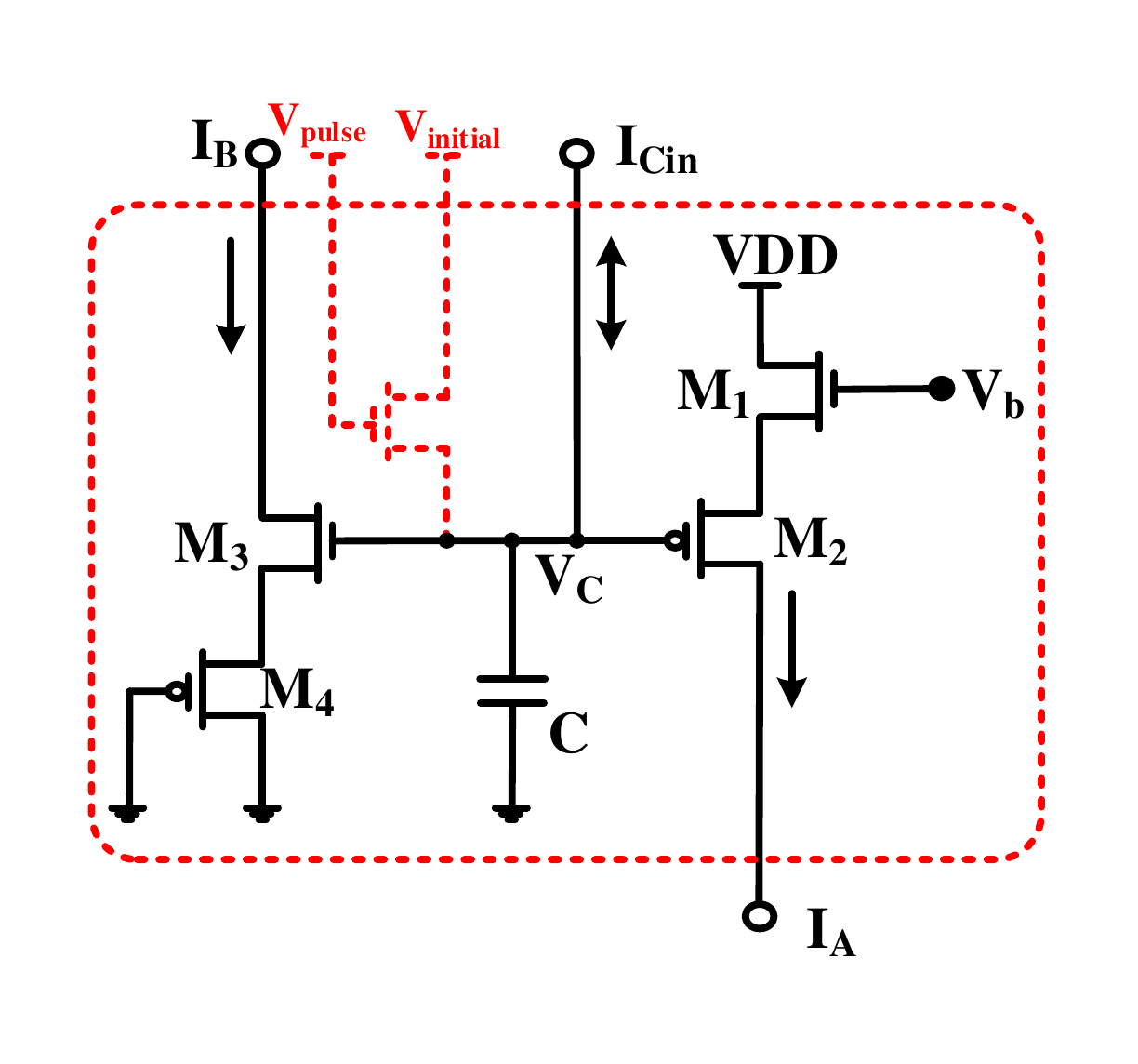}
\vspace{-5pt}
\caption{The ``main core" including the initialization circuit highlighted with red color.}
\label{fig:51_1}
\end{figure}

\begin{equation}\label{eq:51_3}
\dot{I}_A=\overbrace{2k_n(V_{GS}-V_{th})}^{\sqrt {k_nI_A}}\dot{V}_{GS_1}
\end{equation}

\begin{equation}\label{eq:51_4}
\dot{I}_A=\overbrace{2k_p(V_{SG}-V_{th})}^{\sqrt{k_pI_A}}\dot{V}_{SG_2}
\end{equation}

(\ref{eq:51_3}) and (\ref{eq:51_4}) are equal, therefore:
\begin{equation}\label{eq:51_5}
\dot{V}_{SG_2}=\sqrt{\frac{k_n}{k_p}}\dot{V}_{GS_1}=\beta \dot{V}_{GS_1}
\end{equation}
where $\beta=\sqrt{\frac{k_n}{k_p}}$. Similarly, we can derive the following equation for transistors $M_3$ and $M_4$:
\begin{equation}\label{eq:51_6}
\dot{V}_{SG_4}=\sqrt{\frac{k_n}{k_p}}\dot{V}_{GS_3}=\beta \dot{V}_{GS_3}.
\end{equation}
\par The application of Kirchhoff's Voltage Law (KVL) and applying the derivative function show the following relations:
 \begin{equation}\label{eq:51_7}
\dot{V}_{C}=-(\dot{V}_{GS_1}+\dot{V}_{SG_2})
\end{equation}

  \begin{equation}\label{eq:51_8}
\dot{V}_{C}=+(\dot{V}_{GS_3}+\dot{V}_{SG_4})
\end{equation}
 where $V_C$ is the capacitor voltage and $V_b$ the bias voltage which is constant (see Figure \ref{fig:51_1}). Substituting (\ref{eq:51_5}) and (\ref{eq:51_6}) into (\ref{eq:51_7}) and (\ref{eq:51_8}) respectively yields:

 \begin{equation}\label{eq:51_9}
\dot{V}_{C}=-\dot{V}_{GS_1}\cdot(1+\beta)
\end{equation}

\begin{equation}\label{eq:51_10}
\dot{V}_{C}=+\dot{V}_{GS_3}\cdot(1+\beta).
\end{equation}

Setting the current $I_{out}=I_B-I_A$ in Figure \ref{fig:51_1} as the state variable of our system and using (\ref{eq:51_3}) and the corresponding equation for $I_B$, the following relation is derived:

\begin{figure}[t]
\begin{subfigure}{.5\textwidth}
  \centering
\includegraphics[trim = 0in 0in 0in 0in, clip, height=2.3in]{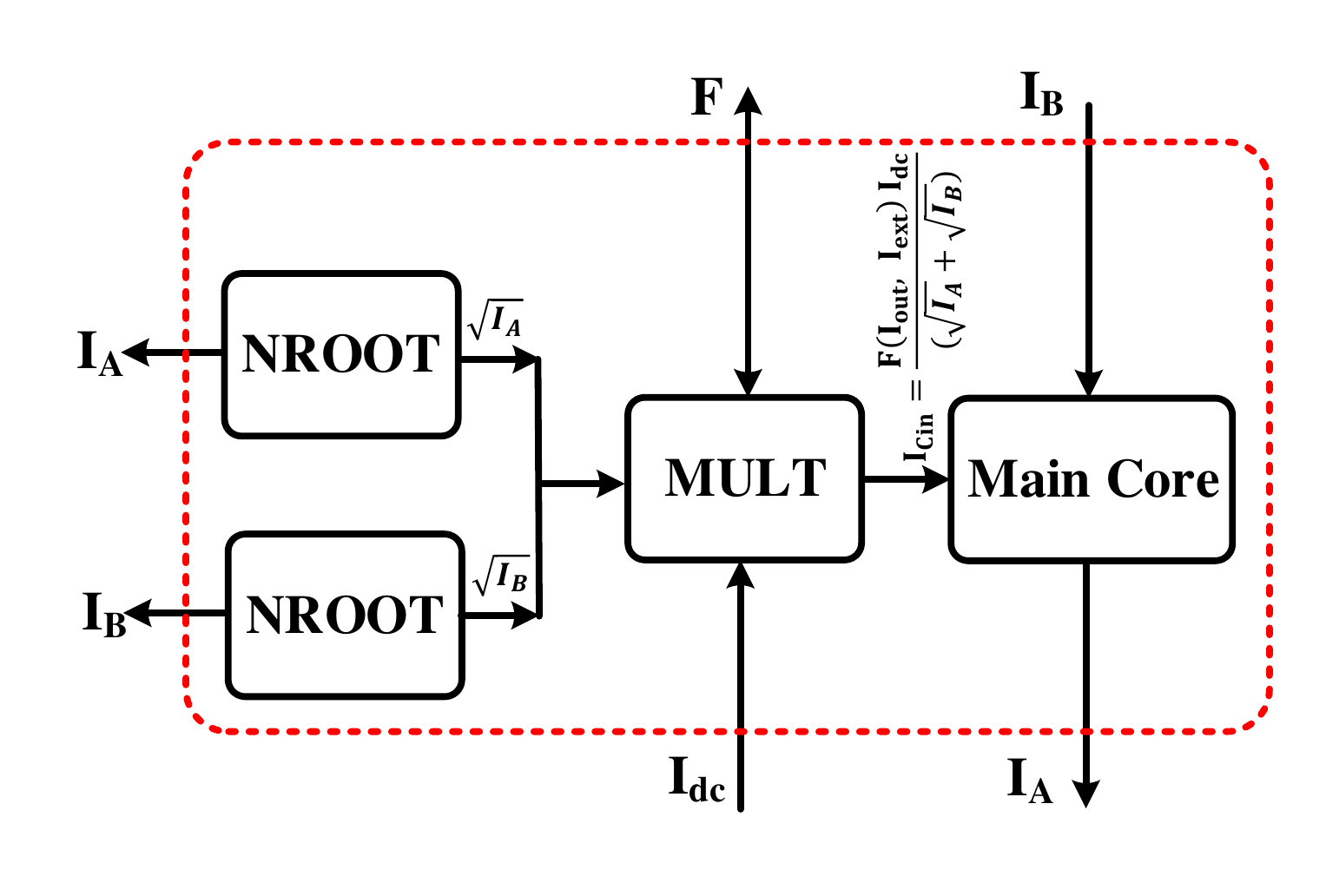}
\label{fig:a}
\end{subfigure}%
\begin{subfigure}{.5\textwidth}
  \centering
\includegraphics[trim = 0in 0in 0in 0in, clip, height=2.5in]{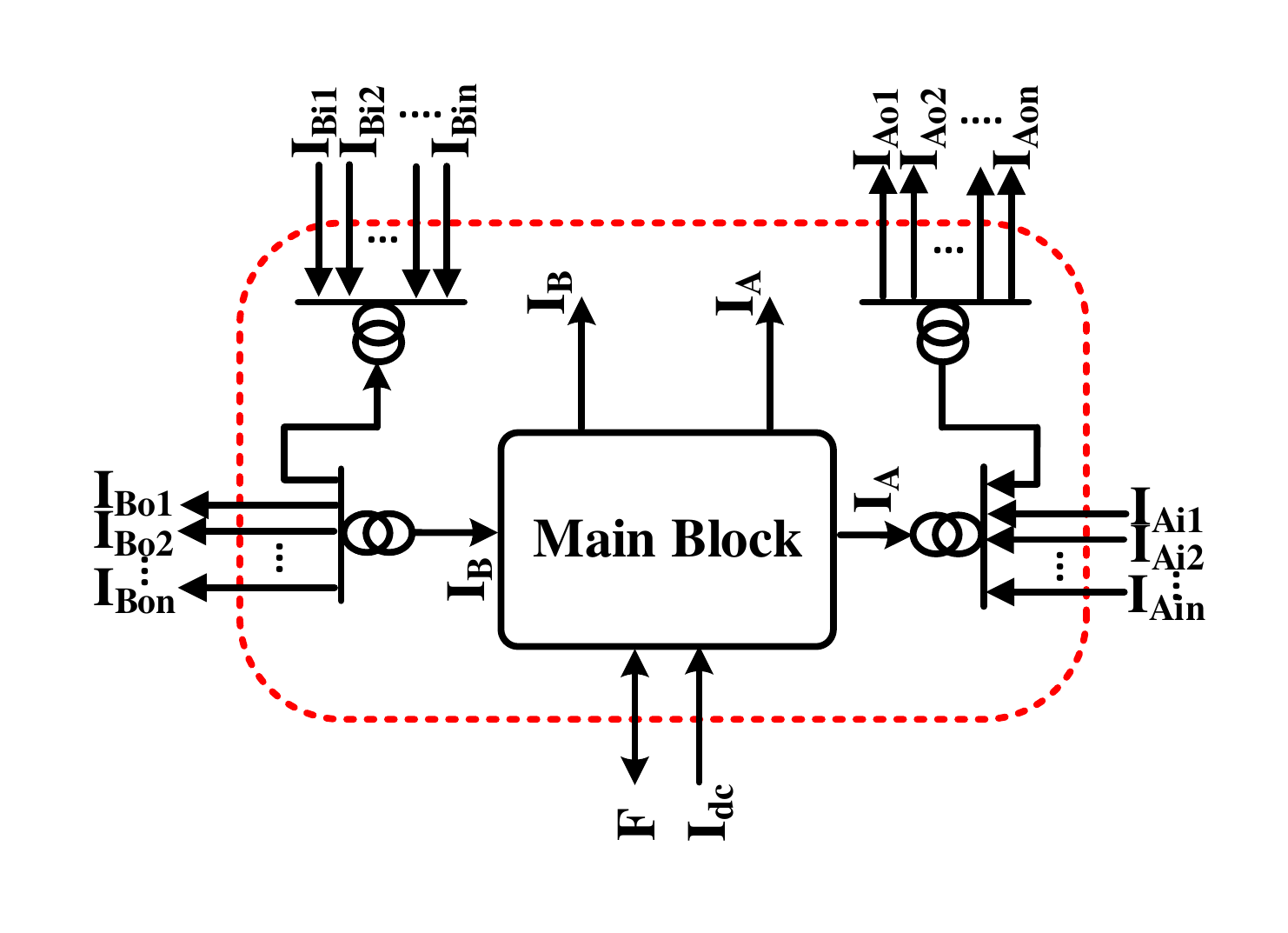}
\label{fig:b}
\end{subfigure}
  \caption{(a) The ``main block" including the main core and two current--mode root square blocks and a bilateral multiplier. (b) The final high speed circuit including the ``main block" with several copied currents (the current mirrors are represented with double circle symbols)}
\label{fig:51_2}
\end{figure}

\begin{equation}\label{eq:51_11}
\dot{I}_{out}=\dot{I}_B-\dot{I}_A=2\sqrt{k_nI_B}\dot{V}_{GS_3}-2\sqrt{k_pI_A}\dot{V}_{GS_1}
\end{equation}

by substituting (\ref{eq:51_9}) and (\ref{eq:51_10}) in (\ref{eq:51_11}):
\begin{equation}\label{eq:51_12}
\dot{I}_{out}=(\sqrt{I_A}+\sqrt{I_B})\cdot \frac{2\sqrt{k_n}\dot{V}_C}{2+\beta}.
\end{equation}
\par Bearing in mind that the capacitor current $I_{Cin}$ can be expressed as $C\dot{V}_C$, relation (\ref{eq:51_12}) yields:
\begin{equation}\label{eq:51_13}
\dot{I}_{out}=(\sqrt{I_A}+\sqrt{I_B})\cdot \frac{2\sqrt{k_n}I_{Cin}}{(2+\beta)C}.
\end{equation}

\par One can show that:
\begin{equation}\label{eq:51_14}
\frac{(2+\beta)C}{2\sqrt{k_n}\cdot I_{dc}}\dot{I}_{out}=\frac{(\sqrt{I_A}+\sqrt{I_B})}{I_{dc}}\cdot I_{Cin}.
\end{equation}
\par Equation (\ref{eq:51_14}) is the main core's relation. In order for a high speed mathematical dynamical system with the following general form to be mapped to (\ref{eq:51_14}):
\begin{equation}\label{eq:51_15}
\tau\dot{I}_{out} =F(I_{out}, I_{ext})
\end{equation}
where $I_{ext}$ and $I_{out}$ are the external and state variable currents, the quantities $\frac{C}{I_{dc}}$ and $I_{Cin}$ must be respectively equal to $\frac{2\tau\sqrt{k_n}}{(2+\beta)}$ and $\frac{F(I_{out}, I_{ext})I_{dc}}{(\sqrt{I_A}+\sqrt{I_B})}$. Note that the ratio value $\frac{C}{I_{dc}}$ can be satisfied with different individual values for $C$ and $I_{dc}$. These values should be chosen appropriately according to practical considerations (see Section V.G). Since $F$ is a bilateral function, in general, it will hold:
\begin{equation}\label{eq:51_16}
I_{Cin}=\overbrace{\frac{F^+(I_A,I_B,I_{ext}^+,I_{ext}^-)I_{dc}}{(\sqrt{I_A}+\sqrt{I_B})}}^{I_{Cin}^+}-\overbrace{\frac{F^-(I_A,I_B,I_{ext}^+,I_{ext}^-)I_{dc}}{(\sqrt{I_A}+\sqrt{I_B})}}^{I_{Cin}^-}
\end{equation}
where $I_{Cin}^+$ and $I_{Cin}^-$ are calculated respectively by a root square block (see Figure \ref{fig:51_2}(a) and $I_{ext}$ is separated to + and -- signals by means of splitter blocks. Note that $I_{dc}$ is a scaling dc current and $\tau$ has dimensions of $second(s)$. Since $I_{Cin}$ can be a complicated nonlinear function in dynamical systems, we need to provide copies of $I_{out}$ or equivalently of $I_A$ and $I_B$ to simplify the systematic computation at the circuit level. Therefore, the higher hierarchical block shown in Figure \ref{fig:51_2}(b) is defined as the NBDS (Nonlinear Bilateral Dynamical System) circuit \cite{16} (see Figure \ref{fig:51_2}(b)) including the main block and associated current mirrors. The form of (\ref{eq:51_15}) is extracted for a 1--D dynamical system and can be extended to $N$ dimensions in a straightforward manner as follows:
\begin{equation}\label{eq:51_17}
\tau_N\dot{I}_{out_N} =F_N(\bar{I}_{out},\bar{I}_{ext})
\end{equation}
where $\frac{C_N}{I_{dc_N}}=\frac{2\tau_N\sqrt{k_n}}{(2+\beta)}$ and $I_{Cin_N}=\frac{F_N(\bar{I}_{out},~\bar{I}_{ext})I_{dc_N}}{(\sqrt{I_{A_N}}+\sqrt{I_{B_N}})}$.

To realized the full potential of \ref{eq:51_17} and implement the aforementioned neuromorphic processing elements, some circuit blocks are required to perform basic operations including multiplication and division. The full description of these blocks can be found in \cite{25}, however, in this paper we only employ them to implement the following case studies. In the following sections, multiple neuromorphic processing elements are implemented in which the feasibility of the above method is shown. 

\begin{figure*}[t]
\vspace{-20pt}
\normalsize
\centering
\includegraphics[trim = 0.1in 0.1in 0.1in 0.1in, clip, width=3.5in]{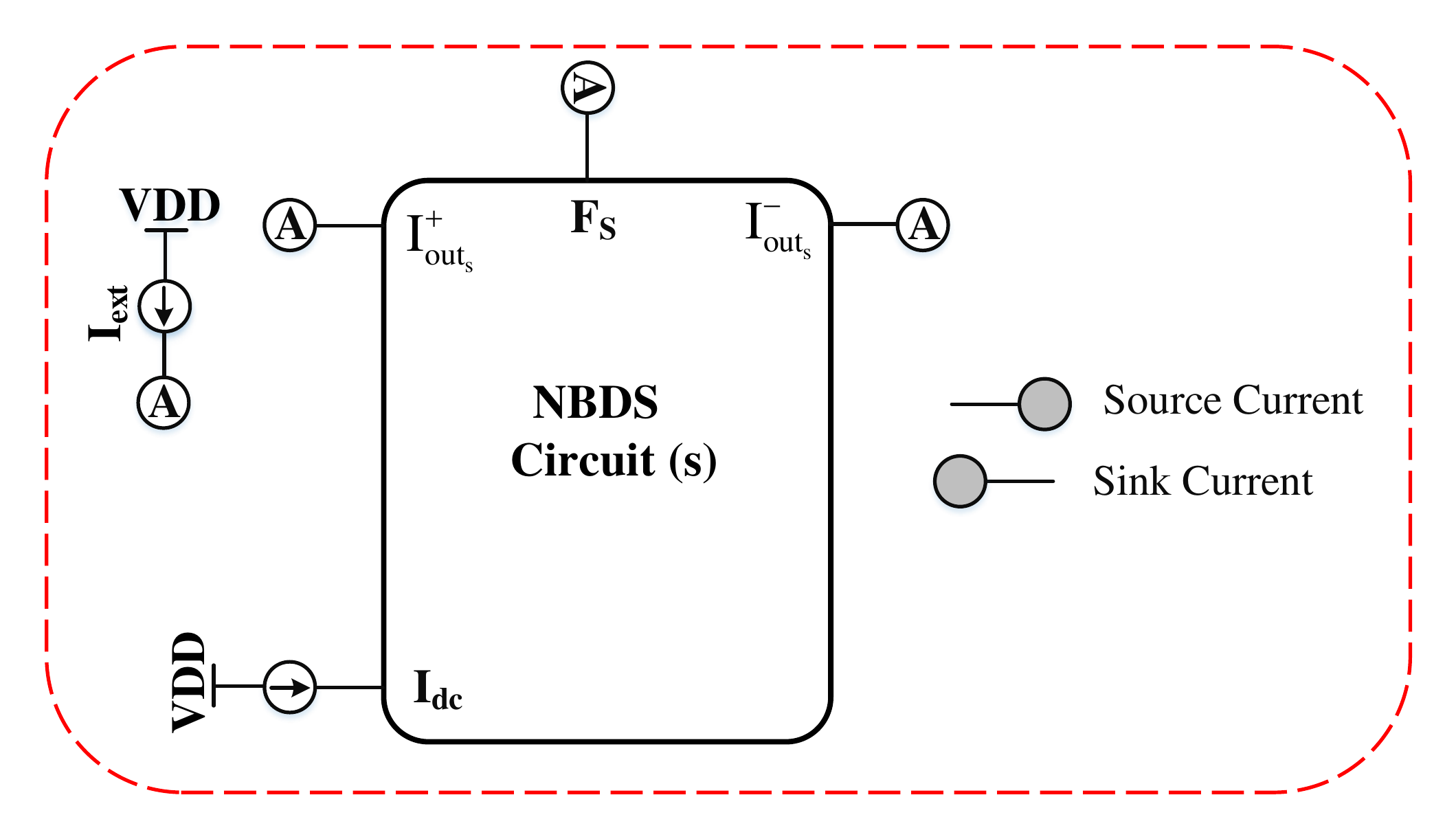}

\caption{A block representation of the total circuit implementing the synapse model.}
\label{fig:51_synapse}
\end{figure*}

\subsection{Synapse model}
Synapses can be represented as low pass filters as the conductance and the capacitance of the post--synaptic cell are connected in parallel, forming a circuit in which the membrane’s capacitance takes time to charge. Therefore they function as low-pass filters attenuating the high frequency components of a pre--synaptic signals. The filtering characteristics including the time scale of a synapse can vary depending upon the membrane properties of the post-synaptic cell. Low pass filters can be generally represented as:
\begin{equation}\label{eq:51_17_5}
 \tau\dot{s}=-s+I_{ext}
\end{equation}
 where $\tau$ defines the response time scale of the synapse. We can start forming the electrical equivalent using (\ref{eq:51_17}):
\begin{equation}\label{eq:51_18}
\begin{cases}
\frac{(2+\beta)C}{2\sqrt{k_n}\cdot I_{dc_s}}\dot{I}_{out_s}=F_s(I_{out_s},I_{ext})
\end{cases}
\end{equation}
where $I_{dc}=I_{dc_s}$, $F_s$ is function given by:
\begin{equation}\label{eq:5_19}
\begin{cases}
F_v(I_{out_s},I_{ext})=-I_{out_s}+I_{ext}\\
\end{cases}
\end{equation}

\par It should be noted that the appropriate size of the capacitor $C$ can be defined based on the relative time scale of the model and $I_{dc}$. It was shown in \cite{16} that there are some trade-offs between the size of the capacitor and $I_{dc}$, in terms of power, area and noise performance. Schematic diagrams for the synapse model is seen in Figure \ref{fig:51_synapse}, including the symbolic representation of the basic electrical blocks introduced in \cite{25}. According to these diagrams, it is observed how the mathematical model is mapped onto the proposed electrical circuit. The schematic contains only one NBDS circuit implementing the only dynamical variable and the necessary connections. As shown in the figure, according to the synapse model parameter, proper bias currents are selected and the correspondence between the biological voltage and electrical current is $V\iff uA$.

\begin{figure*}[t]
\vspace{-20pt}
\normalsize
\centering
\includegraphics[trim = 0.1in 0.1in 0.1in 0.1in, clip, width=6.5in]{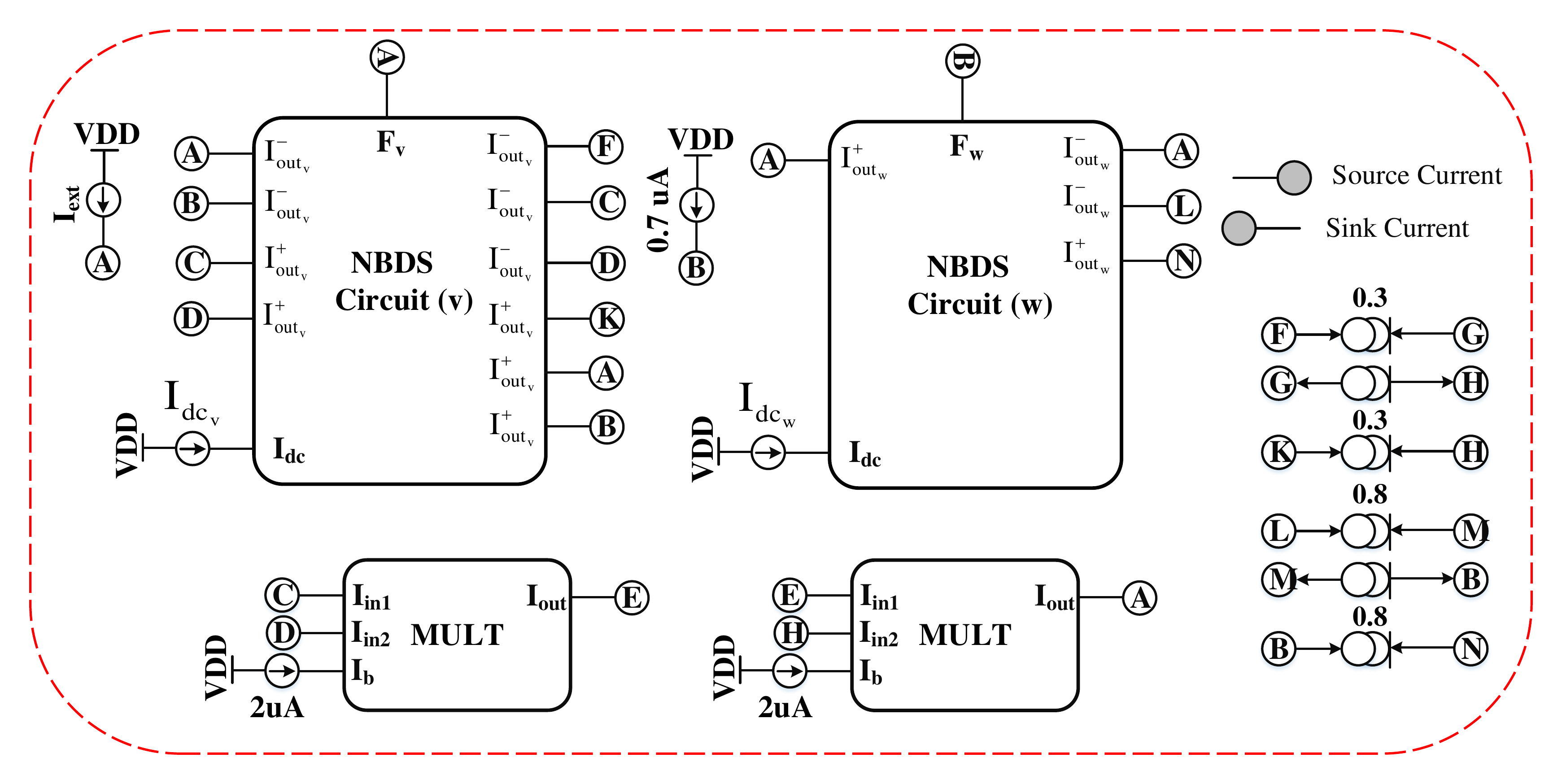}

\caption{A block representation of the total circuit implementing the 2--D FHN neuron model.}
\label{fig:51_neuron}
\end{figure*}

\subsection{Neuron model}
In this section, the application of the method is demonstrated by synthesizing the 2--D nonlinear FitzHugh–Nagumo neuron model. In the FHN neuron model \cite{26} with the following representation:
\begin{equation}\label{eq:51_29_5}
\begin{cases}
 \dot{v}=v-\frac{v^3}{3}-w+I_{ext}\\
  \dot{w}=0.18(v+0.7-0.8w)
 \end{cases}
 \end{equation}
describing the membrane potential's and the recovery variable's velocity. According to this biological dynamical system, we can start forming the electrical equivalent using (\ref{eq:51_17}):
\begin{equation}\label{eq:51_30}
\begin{cases}
\frac{(2+\beta)C}{2\sqrt{k_n}\cdot I_{dc_v}}\dot{I}_{out_v}=F_v(I_{out_v},I_{out_w},I_{ext})\\
\frac{(2+\beta)C}{2\sqrt{k_n}\cdot I_{dc_w}}\dot{I}_{out_w}=F_w(I_{out_v},I_{out_w})
\end{cases}
\end{equation}
where $\tau_v=0.18\tau_w$, and $I_{dc_v}=0.18I_{dc_w}$, $F_v$ and $F_w$ are functions given by:
\begin{equation}\label{eq:5_33}
\begin{cases}
F_v(I_{out_v},I_{out_w},I_{ext})=I_{out_v}-\frac{I_{out_v}^3}{I_bI_x}-I_{out_w}+I_{ext}\\
F_w(I_{out_v},I_{out_w})=(I_{out_v}+I_{c}-\frac{I_dI_{out_w}}{I_x}).
\end{cases}
\end{equation}

\par It should be noted that the appropriate size of the capacitor $C$ can be defined based on the relative time scale of the model. Schematic diagrams for the FHN neuron model is seen in Figure \ref{fig:51_neuron}, including the symbolic representation of the basic blocks introduced in \cite{25}. According to these diagrams, it is observed how the mathematical model is mapped onto the proposed electrical circuit. The schematic contains two NBDS circuits implementing the two dynamical variables, followed by two MULT blocks and current mirrors realizing the dynamical functions. As shown in the figure, according to the neuron model, proper bias currents are selected and the correspondence between the biological voltage and electrical current is $V\iff uA$.

\begin{figure*}[t]
\normalsize
\centering
\includegraphics[trim = 0.1in 0.1in 0.1in 0.1in, clip, width=6.5in]{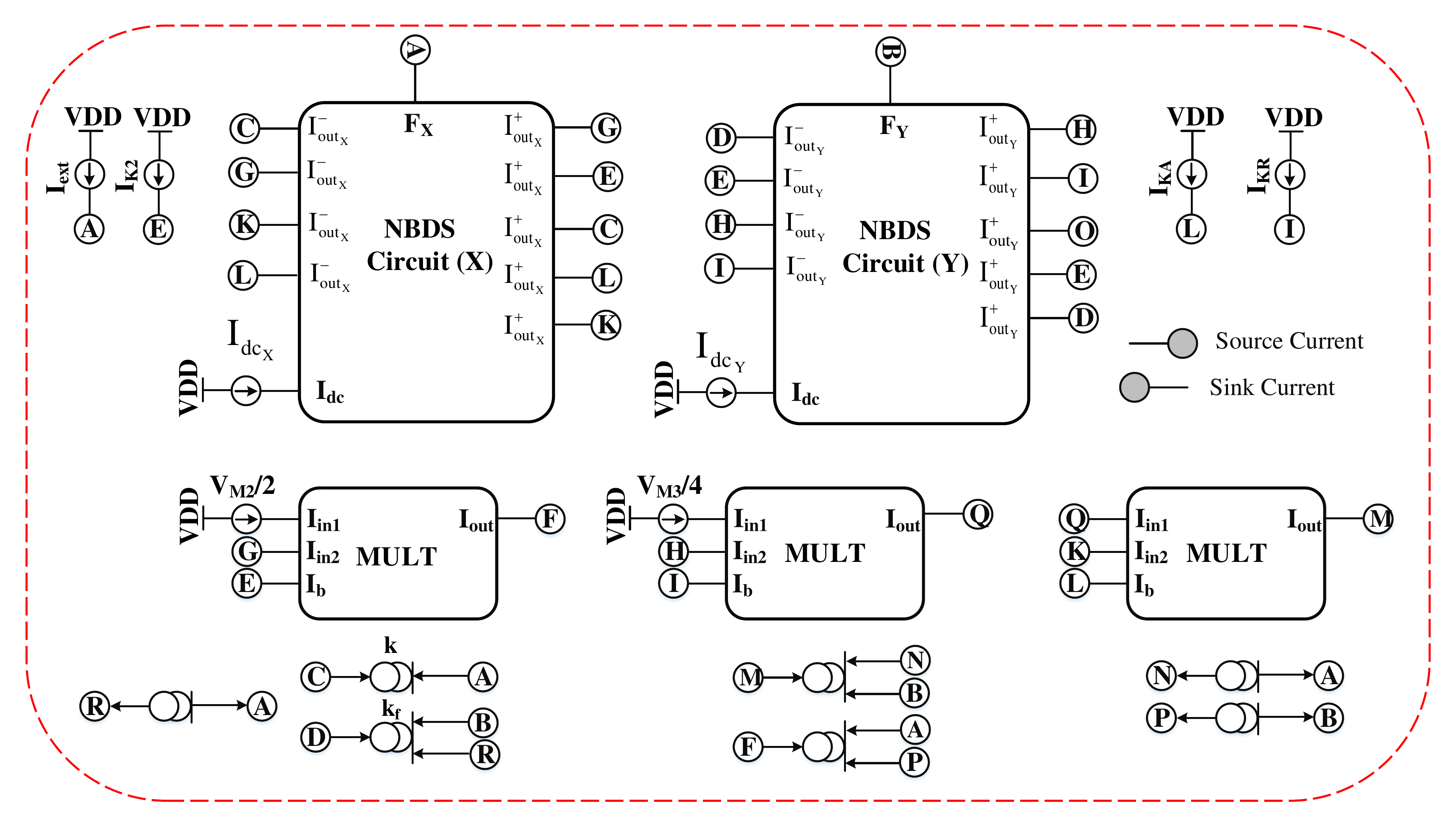}

\caption{A block representation of the total circuit implementing the astrocyte model.}
\label{fig:51_astrocyte}
\end{figure*}

\subsection{Astrocyte model}
Astrocytes are part of the biological neural network that outnumber neurons by over a factor of five in the brain. They occur in the entire central nervous system (CNS) and perform many essential complex functions in the healthy CNS \cite{27}. A reduced and an extended model, which accurately and efficiently describe $Ca^{2+}$  oscillations was presented in \cite{28}. Relying on the fact that the amount of $Ca^{2+}$  released is controlled by the level of stimulus through modulation of the $IP_3$ level and by making the simplification that the level of stimulus-induced, $IP_3$--mediated $Ca^{2+}$  is a model parameter, the following 2--D model was presented:
\begin{equation}\label{eq:51_34}
\begin{cases}
\dot{X}=z_0+z_1\beta - z_2(X) + z_3(X,Y) + k_fY-kX\\
\dot{Y}=z_2(X) - z_3(X,Y) - k_fY
\end{cases}
\end{equation}
where
\begin{equation}\label{eq:51_35}
\begin{cases}
z_2(X)=V_{M_2} \frac{X^n}{K^n_2+X^n}\\
z_3(X,Y)=V_{M_3} \frac{Y^m}{K^m_R+Y^m}\frac{X^p}{K^p_A+X^p}
\end{cases}
\end{equation}
The parameters $V_{M_2}$ , $V_{M_3}$ , $K_2$, $K_R$, $K_A$, $k_f$ and $k$ are the maximum values of $z_2$ and $z_3$, threshold constants for pumping, release and activation and rate constants, respectively. Parameters $n$, $m$, and $p$ define the Hill
coefficients characterizing the pumping, release and activation processes, respectively. Depending on the values of the Hill coefficients, different degrees of cooperativity can be achieved. There are three different cases of Hill coefficients ($m=n=p=1$, $m=n=p=2$ and $m=n=2, p=4$) which in this paper we only explore the $m=n=p=1$ case and leave the rest to the interested readers. The electrical equivalent can be represented as follows: 
\begin{equation}\label{eq:51_30}
\begin{cases}
\frac{(2+\beta)C}{2\sqrt{k_n}\cdot I_{dc_X}}\dot{I}_{out_X}=F_X(I_{out_X},I_{out_Y},I_{ext})\\
\frac{(2+\beta)C}{2\sqrt{k_n}\cdot I_{dc_Y}}\dot{I}_{out_Y}=F_Y(I_{out_X},I_{out_Y})
\end{cases}
\end{equation}
where $I_{ext}=z_0+z_1\beta$, $F_X$ and $F_Y$ are functions given by:
\begin{equation}\label{eq:5_33}
\begin{cases}
F_X(I_{out_X},I_{out_Y},I_{ext})=-\frac{I_{V_{M_2}}I_{out_X}}{I_{K_2}+I_{out_X}}+\frac{I_{V_{M_3}}I_{out_Y}}{I_{K_R}+I_{out_Y}}\frac{I_{out_X}}{I_{K_A}+I_{out_X}}-kI_{out_X}+k_fI_{out_Y}+I_{ext}\\
F_Y(I_{out_X},I_{out_Y})=\frac{I_{V_{M_2}}I_{out_X}}{I_{K_2}+I_{out_X}}-\frac{I_{V_{M_3}}I_{out_Y}}{I_{K_R}+I_{out_Y}}\frac{I_{out_X}}{I_{K_A}+I_{out_X}}-k_fI_{out_Y}.
\end{cases}
\end{equation}
 
\par It should be noted that the appropriate size of the capacitor $C$ can be defined based on the relative time scale of the model. Schematic diagrams for the Astrocyte $Ca^{2+}$ oscillations model is seen in Figure \ref{fig:51_astrocyte}, including the symbolic representation of the basic blocks introduced in \cite{25}. According to these diagrams, it is observed how the mathematical model is mapped onto the proposed electrical circuit. The schematic contains two NBDS circuits implementing the two dynamical variables, followed by three MULT blocks and current mirrors realizing the dynamical functions. As shown in the figure, according to the neuron model, proper bias currents are selected and the correspondence between the biological voltage and electrical current is $V\iff uA$.

\section{Discussion}
\par In this paper, we demonstrated a unifying tutorial which synthesizes biological neuromorphic models into analog circuitry with the minimum effort. The validity of the approach was shown by implementing multiple neuromorphic processing elements including synapse, neuron and astrocyte models. As mentioned above there are multiple design trade-off aspects (including but not limited to the capacitor sizes and bias currents ($I_{dc}$)) that interested readers should be aware of in order to design a power efficient circuity. 

\bibliographystyle{unsrt}	
\bibliography{paper}

\end{document}